\documentclass[12pt,a4paper]{article}

\usepackage{graphicx}  % To import .pdf files
\usepackage[margins]{trackchanges}

\usepackage[backend=bibtex,style=authoryear,natbib=true]{biblatex}
\usepackage[autostyle=true]{csquotes}
\usepackage{import}
\usepackage{amsmath}
\usepackage[colorlinks,linkcolor=blue,citecolor=blue,urlcolor=blue,bookmarks=false,hypertexnames=true]{hyperref}
\usepackage{float}
\usepackage{tabularx}
\usepackage{makecell}
\usepackage{mathtools}
\usepackage{amsfonts}
\usepackage{bm}
\usepackage{booktabs}
\usepackage{multirow}
\usepackage{threeparttable}
\usepackage{array}
\usepackage{changepage}
\usepackage{graphicx} 
\newcommand{\Thickrule}{\noalign{\global\heavyrulewidth=1pt}\toprule}

\usepackage{multirow}
\usepackage{booktabs}
\makeatletter
\newsavebox{\@tabnotebox}

\providecommand\tnote{}

\makeatother

\usepackage{xargs}
\usepackage[pdftex,dvipsnames]{xcolor}
\usepackage[colorinlistoftodos,prependcaption,textsize=tiny]{todonotes}
\newcommandx{\unsure}[2][1=]{\todo[linecolor=red,backgroundcolor=red!25,bordercolor=red,#1]{#2}}
\newcommandx{\changeA}[2][1=]{\todo[linecolor=blue,backgroundcolor=blue!25,bordercolor=blue,#1]{#2}}
\newcommandx{\info}[2][1=]{\todo[size=large,linecolor=OliveGreen,backgroundcolor=OliveGreen!25,bordercolor=OliveGreen,#1]{#2}}
\newcommandx{\improvement}[2][1=]{\todo[linecolor=Plum,backgroundcolor=Plum!25,bordercolor=Plum,#1]{#2}}
\newcommandx{\thiswillnotshow}[2][1=]{\todo[disable,#1]{#2}}

\RequirePackage[normalem]{ulem} %DIF PREAMBLE
\RequirePackage{color}\definecolor{RED}{rgb}{1,0,0}\definecolor{BLUE}{rgb}{0,0,1} %DIF PREAMBLE
\begin{document}

\begin{Large}
\begin{center}
\textbf{Batch Aggregation: An Approach to Enhance Text Classification with Correlated Augmented Data}
\end{center}
\end{Large}

\begin{center}
\textbf{Charco Hui, Yalu Wen} \\
The University of Auckland, Department of Statistics \\
\smallskip
\small{\{charco.hui, y.wen\}@auckland.ac.nz}
\end{center}

\begin{center}
\today
\end{center}

\begin{abstract}
Natural language processing  models often face challenges due to limited labeled data, especially in domain specific areas, e.g., clinical trials. To overcome this, text augmentation techniques are commonly used to increases sample size by transforming the original input data into artificial ones with the label preserved. However, traditional text classification methods ignores the relationship between augmented texts and treats them as independent samples which may introduce classification error. Therefore, we propose a novel approach called `Batch Aggregation' (BAGG) which explicitly models the dependence of text inputs generated through augmentation by incorporating an additional layer that aggregates results from correlated texts. Through studying multiple benchmark data sets across different domains, we found that BAGG can improve classification accuracy. We also found that the increase of performance with BAGG is more obvious in domain specific data sets, with accuracy improvements of up to 10–29\%. Through the analysis of benchmark data, the proposed method addresses limitations of traditional techniques and improves robustness in text classification tasks. Our result demonstrates that BAGG offers more robust results and outperforms traditional approaches when training data is limited.
\end{abstract}
\section{Introduction} \label{sec:intro}

Natural language processing (NLP) is a fast growing field that has made significant impacts. For example, e-commerce platforms such as Amazon, eBay and Shopify have already taken advantage of NLP to understand their customers’ behavior, predict customer purchases, and learn valuable insights for decision making. Moreover, in the education sector, NLP can be utilized to generate feedback and assist students in writing and reading comprehension, or provide useful tools to help teachers reduce time-consuming work such as marking assessments. In addition, NLP has also been used in the medical field, such as comprehending medical and clinical trials reports for disease classification \citep{ct-disease} and event detection \citep{ct-event}. The advances in NLP in medicine are able to reduce repetitive manual work, and provide useful insights to practitioners or researchers to increase productivity and efficiency. 

Text classification is one of the most active area of research in the field of NLP due to its wide range of applications, such as sentiment analysis, topic classification and spam filtering. The most common methods used for text classification in the field of NLP includes Recurrent Neural Network (RNNs), Long Short-Term Memory \citep[LSTM;][]{lstm}, Seq2seq \citep{seq2seq} and Bidirectional Encoder Representations from Transformers \citep[BERT;][]{bert}. RNNs are a type of neural network where inputs are passed into the model sequentially; this characteristic matches the nature of human language, therefore its well suited for handling text data. However, since inputs are passed into the model sequentially, it will require all previous hidden states to compute an output causing it to be slow at inference. Moreover, vanishing gradients and information loss are common due to the combination of hidden states. LSTM is a variant of RNN that was designed to address issues of information loss by introducing a memory block to store information between hidden states but the problem of heavy computation still remains. 

On the other hand, Transformers \citep{transformers}, by utilizing its multi-head attention mechanism, can achieve parallelization at processing input while maintain bidirectional properties which overcomes the problem of vanishing gradients and slow computation in traditional methods. Furthermore, BERT utilized the Transformers architecture and introduced masked language modeling and next sentence prediction, a pre-training approach to improve language comprehension substantially. These pre-trained models can then be fine-tuned for various downstream tasks by such as question answering and text classification to achieve state-of-the-art performance. For example, \citet{finbert} used BERT to comprehend financial texts then further fine-tuned the model for financial sentiment classification. In addition, \citet{hatespeech} and \citet{clinicalbert} also leveraged BERT for classifying hate speech and clinical trial notes. The underlying idea is to utilize an existing BERT model and fine-tune it with additional data, adding one or more output or activation layers on top of the BERT model to transform the output embeddings into class probabilities. 

While these methods have achieved the state-of-the-art performance in many open domains where a large amount of training data is readily available, their applicability in domain-specific area (e.g., clinical trials, finance, medicine) has been limited primarily due to the relatively small training sample. Indeed, to gain insight with text classification for domain specific areas, the common approach is to acquire a substantial amount of annotated data in the desirable domain \citep{clinicalbert, domain2, domain3} or in general improve existing model architectures and train it from scratch \citep{roberta, albert, electra}. However, improving model architecture often requires a significant amount of computational resources \citep{compute-resource1, compute-resource2, compute-resource3} and retrieving high-quality labeled data is a labor-intensive and time-consuming job that heavily relies on the availability of existing domain experts. Therefore, there is an urgent need to develop a powerful NLP model for text classification when training data is limited. 

Data augmentation, a method which increases the sample size by transforming the original input data into multiple artificial ones with the label preserved, is a commonly used approach for NLP when the training data is limited. With augmented data, deep learning models can benefit from the increase in sample size and diversity between texts to boost model performance and reduce over-fitting. Augmentation methods, such as rotating, flipping and distorting, have already been widely used in the field of computer vision. However, in the field of NLP, the adoption of text augmentation has only been popularized in recent years primarily due to the lack of performant language models to comprehend the complexity of human languages. Similar to image augmentation, the goal of text augmentation is to increase sample size and diversity to the training data so that the text model can be generalized to unseen data. However, altering text such as inserting a random word may impact the original meaning of a sentence significantly, thus applying text augmentation needs to be handled with caution to prevent generating misleading information. 

The most straightforward approach for text augmentation in NLP is to apply augmentation methods on words or parts of words (token level) such as synonym \citep{synonym-replacement} and entity \citep{entity-replacement} replacement and paraphrasing \citep{paraphrasing}. \citet{eda} also proposed Easy Data Augmentation (EDA) which consists of a combination of four simple but powerful operations: (i) synonym replacement, (ii) random word insertions, (iii) swaps and (iv) deletions. While EDA is a quick method to introduce diversity to the data, its simplicity may result in loss of key information and its limited in altering sentence structure due to its simplicity, leading in less diversity between texts. Beyond token level, a variety of methods has been investigated for augmentation at sentence level such as sentences insertion \citep{nlpaug}. Additionally, there exists approaches such as summarization and paraphrasing methods \citep{para-sum1, para-sum2, nlpaug} which reconstruct or shorten texts while preserving their original meaning. Typically, there are two types of summarizations: extractive and abstractive. The first aims to extract sentences or paragraphs to form the summary, where these selected texts should be able to represent the core meaning of the original text. The second type, abstractive summarization, takes a more flexible approach by generating new sentences that may or may not be present in the original text while keeping the underlying meanings of the original text. As for paraphrasing, it is similar to abstractive summarization where it is able to generate new sentences but generally does not shorten the original text by a significant amount.

Another notable method for text augmentation in NLP is to utilize machine translation \citep{translation1, translation2, translation3}, which involves translating a sentence to another language then back to its original language. In fact, the method of translation shares similarities with paraphrasing, as they both aim to reconstruct sentences while preserving its original meaning since different languages have different sentence structures thus translating to and from another language should be able to achieve the same effect. The two common translation approaches are to use online services (e.g., Google Translate\footnote[1]{https://translate.google.com/}, DeepL\footnote[2]{https://www.deepl.com/}) or to utilize a translation model powered by neural networks (e.g., OPUS-MT \citep{opus-mt}, Marin \citep{marin}). In contrast to EDA, translation-based augmentation methods can alter sentence structure while preserving the original meaning, but these methods are often computation heavy and is usually not cost-effective. Furthermore, the quality of translation with online services (e.g., Google Translate, DeepL) cannot be controlled, since these services are not open-sourced. As for neural network based translation models (i.e., OPUS-MT, Marin), although the translation quality can be controlled, the output’s embeddings may be similar to each other, and if a transformers based model is used on these text for prediction, the diversity in each augmented text may be low, embedding-wise.

\subsection{The Proposed BAGG Technique} \label{subsec:intro-ba}

While text augmentation has proven to improve model performance \citep{aug-improve1, aug-improve2, aug-improve3, aug-improve4, aug-improve5}, existing methods treat augmented text as if they were independent samples which unintentionally introduce correlations among inputs, and can affect the downstream tasks such as text classification. The classification errors coming from the inputs that are generated from the same text are no longer independent of each other, and the expected classification prediction errors are not guaranteed to converge to the true error of the model. This could lead to over-fitting and biased estimate in the predicted probability for each class. Indeed, it has well been established treating correlated inputs as if they were independent can lead to the underestimation of variance in the estimators \citep{corr-input, corr-input2, corr-input3}. For example, the stochastic gradient descent and its variants can be biased estimator of the full gradient \citep{corr-sdg} when empirical data distribution does not correspond to the data generating distribution.

Although there has been some study in transferring knowledge from traditional method theories to neural networks to handle correlated inputs, for example, Mixed Model Neural Network \citep{menn}. However, to the best of our knowledge, there are none that focuses on correlated augmented text inputs for neural networks in the NLP field. Solving this problem is crucial since it can affect the reliability and robustness of the classification model. Therefore, to address these issues, we utilized the idea in analyzing clustered data and developed a new method which we call \textit{`Batch Aggregation'}, where the dependence of the inputs generated from augmentations are explicitly considered and modelled. Specifically, we have designed the batch aggregation layer to first summarize results from correlated inputs and then make the final text classification. Loss is then calculated for each observation rather than each text input which reduces the impact of correlation among augmented text. We demonstrate here that BAGG has substantial advantage over competing methods. 

In the following sections, we first describe the methodology in Section 2. Section 3 introduces the benchmark data and presents the results. Finally, Section 4 discusses our findings, their implications and some potential future work.
\section{Batch Aggregation Methodology} \label{sec:method}

Traditional classification methods treat augmented texts as if they were independent new observations, and they completely ignore the correlations among augmented texts as well as the contextual relationships between augmented texts and their original forms. This can lead to bias estimates, sub-optimal classification performance and over-fitting which may cause the model to focus on memorizing the training data instead of the general pattern \citep{alphago}. To combat for these issues, our BAGG method models correlations among augmented texts explicitly.

Let $\bm X_i=\{X_{i0}, \bm X^{T}_{a,i}\}=\{X_{i0}, X_{i1}, \cdots, X_{in_i}\}$, where $X_{i0}$ is the original $i$th observation and $\bm X_{a,i}$ is the vector of its $n_i$ augmented text. Let $h(x;\bm \theta)=g(f(x;\bm \theta_1);\bm \theta_2)$ represent the classification model, where $\bm \theta = (\bm \theta_1^T, \bm \theta_2^T)^T$, $f(\cdot; \bm \theta_1)$ is the output from the layers that capture the structure of the input sentence, and $g(\cdot; \bm \theta_2)$ is output for the final classification problem. For simplicity and without loss of generality, we focused on the BERT classification model that has achieved the state-of-the-art performance in NLP, and thus $f(\cdot; \bm \theta_1)$ is the pre-trained BERT model and $\bm Z = g(\cdot; \bm \theta_2)$ is the layer associated with the prediction task, where $\bm Z$ represents the final model output. For traditional BERT classification model with augmented texts, the loss function is defined as 

\begin{align}
\label{eqn:bertloss}
J(\bm \theta) &= E_{(\bm x, y) \sim \widehat{p}_{data}}L(h(\bm x; \bm \theta),y) \nonumber \\
& = \frac{1}{n}\sum_{i=0}^n\frac{1}{n_i}\sum_{j=0}^{n_i}l(h(x_{ij};\bm \theta), y_i),
\end{align}
where $n$ is the sample size and $\widehat{p}_{data}$ is the empirical distribution. Since the final task is a classification problem, cross-entropy is usually used to measure the loss. 

The most common way to optimize parameters for \eqref{eqn:bertloss} is through the use of minibatch and the key assumption for obtaining an unbiased estimate of the expected gradient is that the random set of samples are independent of each other and drawn from the same data generating distributions. It is well-known that the minibatch stochastic gradient descent is consistent with the gradient of $E_{(\bm x, y) \sim {p}_{data}}L(h(\bm x; \bm \theta),y)$, where the expectation is taken over the data generating distribution ${p}_{data}$ rather than its empirical estimates $\hat{p}_{data}$. As $\bm{x}_{a,i}$ are augmented from the same input $x_{i0}$, the observation pair $(x_{ij}, y_i)$ and $(x_{ij'}, y_i)$ is correlated and thus $\text{cor}(l_{ij}, l_{ij'}) \neq 0$ for $j \neq j' $. Thus the model may produce biased estimates of the full gradient since the samples are correlated \citep{corr-sdg}. Additionally, despite augmentations provide additional variations in the text that can enhance the learning of text representations, the efficient sample size with $n_i$ augmentations for the $i$th input is still less than $n_i+1$. Thus directly optimizing \eqref{eqn:bertloss} and treating augmented text as if they were observed samples can result in a biased estimate in the model parameters and affects the estimated probability for the classification.

\begin{figure}[h]
	\centering
	\includegraphics[scale=0.35]{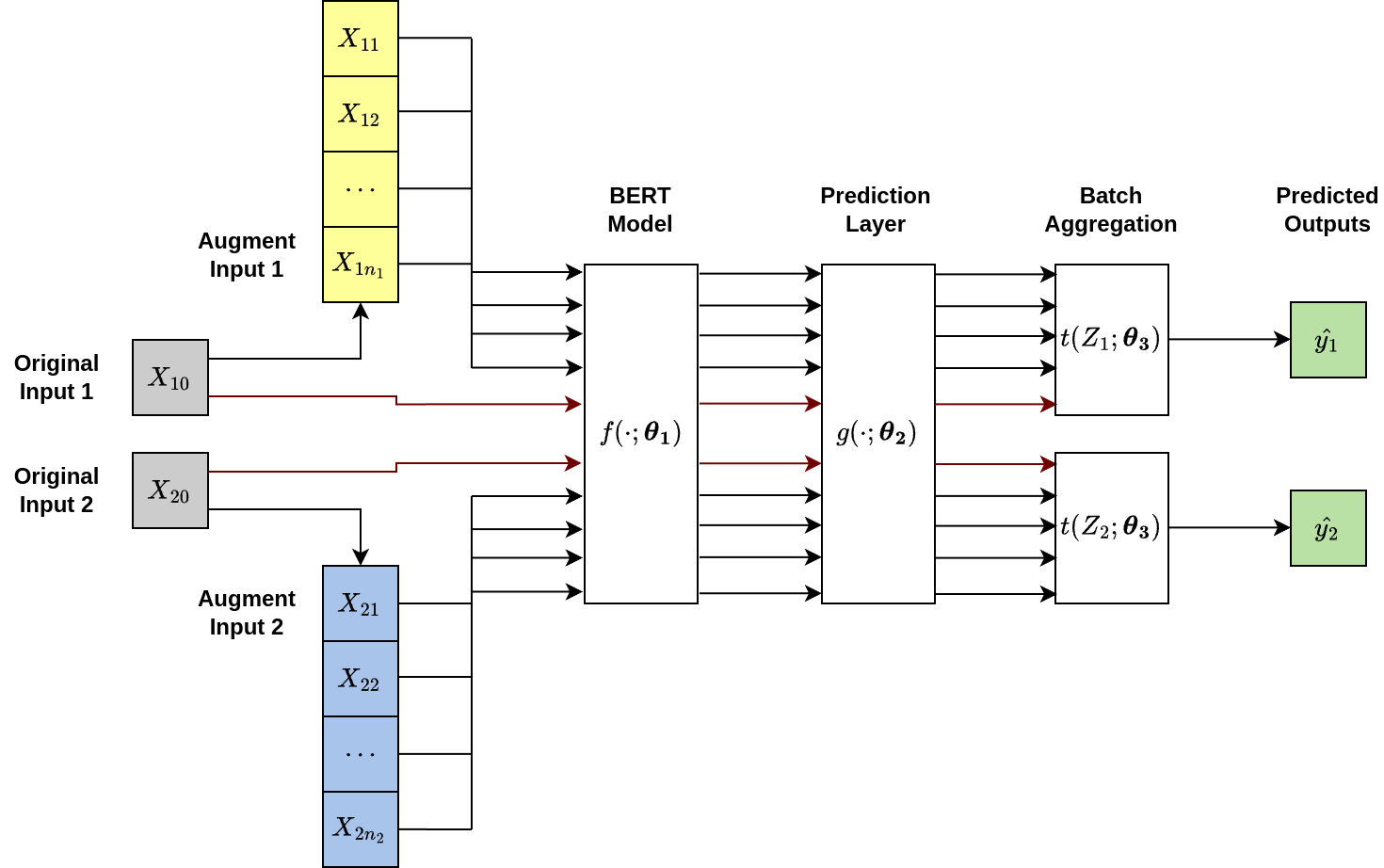}
	\caption{Batch Aggregation (BAGG) with two inputs}
	\label{fig:batch-aggregation}
\end{figure}

For data that has a nested structure (e.g., clustered data), hierarchical modeling or mixed-effects modeling are one of the mostly widely used technique, where random effects are employed to account for correlations. Following a similar idea used in these models, we view our augmented data as a type of clustered data, and propose the sampling unit to be each observation $(\bm X_i, y_i)$ rather than each text input (i.e., $(X_{ij}, y_i)$) that is used in standard BERT models. We still use the function $f(\cdot; \bm \theta_1)$ to learn the language structures from the original input $x_{i0}$ and their variations obtained from augmentation (i.e., $\bm x_a'$), but we proposed to introduce a pooling layer to (denoted using function $t(\cdot; \bm \theta_3)$) to first summarize the information from augmented texts that are generated from the same input and then make the prediction based on the summarized information using $g(\cdot;\bm \theta_2)$, which we denote as $\bm Z$. Therefore, the predicted outcome for the $i$ observation is $h'(\bm x_i; \bm \theta')=t \left( g(f(x_{i};\bm \theta_1);\bm \theta_2);\bm \theta_3 \right)$. Note that the function $t(\cdot;\bm \theta_3)$ associated with our proposed pooling layer can be chosen to reflect the relative importance of each augmentation. In our study, we choose to treat all augmented texts the same and used the average function to summarize their information. The general rationale of our proposed algorithm is depicted in \autoref{fig:batch-aggregation}, and the overall loss function of our proposed algorithm is

\begin{equation}
\label{eqn:baloss}
J'(\bm \theta') = \frac{1}{n}\sum_i^n\l(\bm x_{i}, y_i)=\frac{1}{n}\sum_i^n\l\left(h'(\bm x_{i};\bm \theta'), y_i\right).
\end{equation}

By conditioning on the input text in the pooling layer, we have changed sampling unit to each independent observation $\bm x_i$, which is sensible to assume to come from the same data generating distribution. As such, it is possible to obtain an unbiased estimate of the gradient by using the average gradient obtained on a minibatch with $n_m$ independent samples. The gradient in \eqref{eqn:baloss} that is the expectation of loss taken over finite training set is expected to be consistent with that where expectation is taken over the data generating distribution. With our proposed pooling layer, the loss is calculated for each input rather than for each augmented text $x_{ij}$, and the correlation between $l(\bm x_i, y_i)$ and $l(\bm x_j, y_j)$ is expected to be 0 when $i \neq j$. We avoid the estimation of the joint distribution of augmented texts and the correlation between $l(h(x_{ij};\bm \theta), y_i)$ and  $l(h(x_{ij'};\bm \theta), y_i)$ that is hard to obtain in practice. With pooling layer incorporated, there are no correlations between individual losses (i.e., $l_i$), and the traditional theory and algorithms in neural network apply. 

While multiple augmentation methods such as EDA and back-translation have been proposed in the existing literature, none is consistently superior. A trial-and-error approach is commonly adopted to search for an optimal augmentation method for the problem at hand. To address this challenge, we propose a novel solution that considers augmentations from different methods (e.g., EDA, OPUS-MT, and Google Translate). Specifically, borrowing from ensemble learning and blending techniques, instead of relying on a single augmentation method to derive $\bm X^{T}_{a,i}$, we combine multiple augmentation methods to generate a larger and more diverse input $X_i$. Suppose a total of $k$ augmentation methods are considered. Let $a_j, j \in \{1, 2, \cdots, k\}$ represent the $j$th augmentation method and $\bm X^{T}_{i, a_j} = \{X_{i,a_j,1}, X_{i,a_j,2}, \dots, X_{i,a_j,n_j}\}$ be the corresponding $n_j$ augmented texts for the input $x_{i0}$. The final set of input based on the $i$th input text is $X_i=\{X_{i0}, \bm X^{T}_{i,a_1}, \bm X^{T}_{i,a_2}, \cdots \bm X^{T}_{i,a_k}\}$, where each $\bm X^{T}_{i,a_j}$ captures a different representation of our original input $X_{i0}$ through apply different augmentation methods. Instead of applying one augmentation method on $X_{i0}$, $k$ methods can be used, for example $a_1$ may be utilizing EDA and $a_2$ may be back translation. As a result we obtain a new set of input $\bm X_i=\{X_{i0}, \bm X^{T}_{a_1,i}, \bm X^{T}_{a_2,i}, \cdots \bm X^{T}_{a_k,i}\}$ where each $\bm X^{T}_{a_k,i} = \{X_{a_k,i1}, X_{a_k,i2}, \dots, X_{a_k,in_k}\}$ captures a different representation of our original input $X_{i0}$ through apply different augmentation methods. Since all inputs are still derived from $X_{i0}$, it is still sensible to assume that all augmented inputs $\bm X_{a_k,i}$ are correlated even though each augmentation methods may vary, and correlations are likely higher between methods compared to across methods. However, accounting for the variability introduced by different augmentation methods, a more robust model may result. By allowing the model to learn the characteristics of each augmentation method it can attempt to understand the contextual representations through different perspectives. This consideration should further improve the model's ability to extract insights from varied inputs, thus improving its generalization capabilities.

\begin{figure}[h]
	\centering
	\includegraphics[scale=0.35]{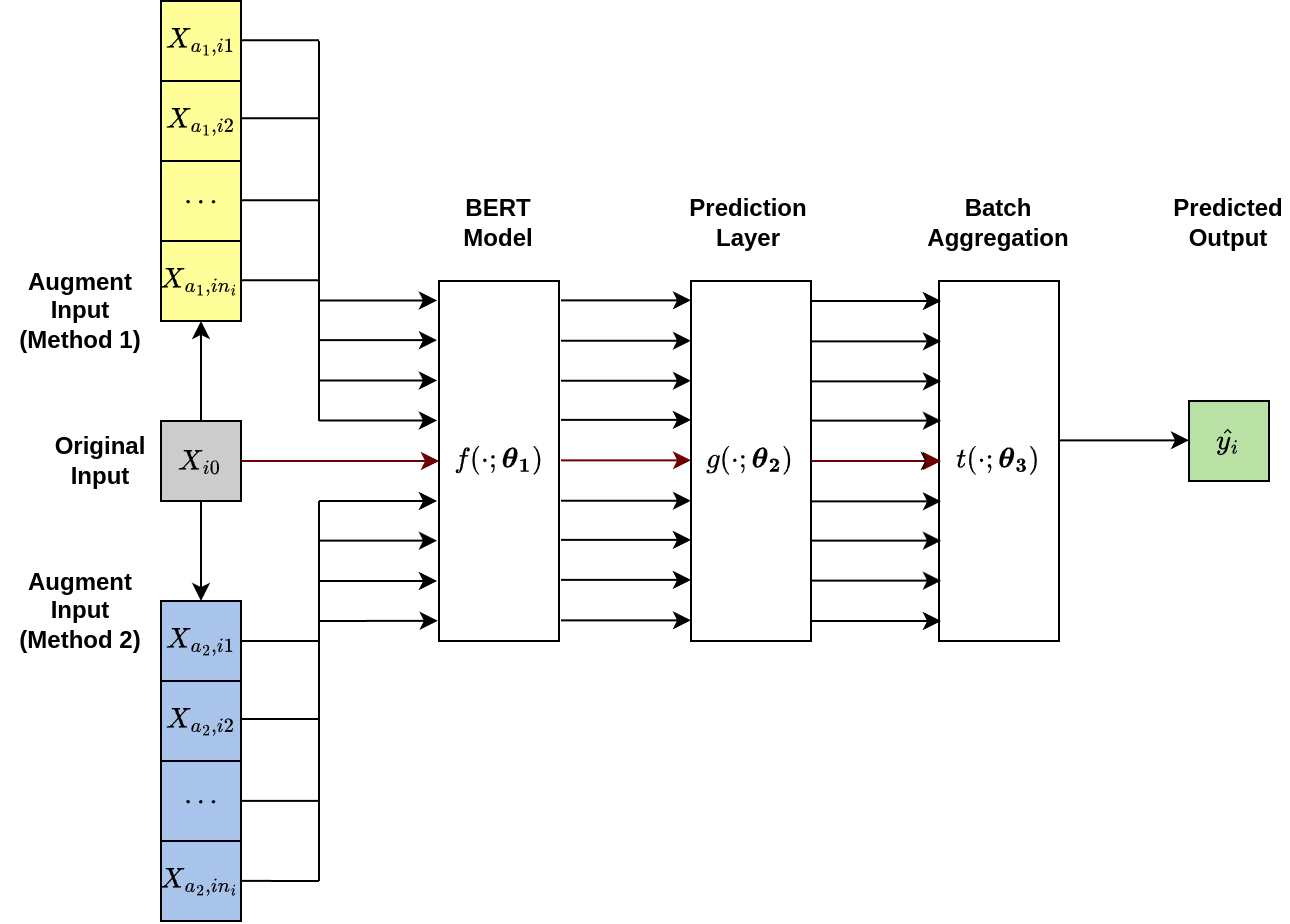}
	\caption{Batch Aggregation with two augmentation methods}
	\label{fig:batch-aggregation-combined}
\end{figure}

In our study, we chose to use EDA and back-translation with OPUS-MT and Google Translate, since these methods are able to cover multiple augmentation variations including, synonym replacement, random word insertions, swaps, deletions and paraphrasing with both cloud providers and neural networks. Summarization was not considered since a portion of our input text were too short. While we could construct an additional pooling layer to first aggregate prediction from each augmentation method and then use another layer to further aggregate the results from each augmentation such as utilizing two batch aggregation layers, supplementary \hyperref[fig:batch-aggregation-2pool]{Figure~\ref{fig:batch-aggregation-2pool}}), we consider this unnecessary. This is mainly because the general framework that we proposed cover this special condition, unless the functions associated with pooling are different. 
The general idea of the algorithm is shown in \autoref{fig:batch-aggregation-combined}. Here, the proposed extension to BAGG builds upon the original approach, thus the model architecture and loss function will remain the same as in \eqref{eqn:baloss}. 

\clearpage
\section{Analysis of Benchmark Data} \label{sec:data}

To examine the performance of our proposed BAGG technique, we conducted benchmark studies with three publicly available data sets: the Amazon Customer Reviews data set\footnote[3]{https://huggingface.co/datasets/amazon\_reviews\_multi/}, the 20 newsgroups text data set\footnote[4]{https://huggingface.co/datasets/rungalileo/20\_Newsgroups\_Fixed/} and the LitCovid data set\footnote[5]{https://huggingface.co/datasets/KushT/LitCovid\_BioCreative/}. These three were chosen due to their differences in domain and variability in inputs and outputs, text length and number of categories. As the sample size and the number of categories are important factors that affect classification, we considered their impacts by randomly sampling observations from the above three data sets, where a sample size of 100 and 200 and the number of categories of 8 and 12 are considered. Here, large sample sizes are not considered since augmentation methods are designed to deal with issues of small data sets and these algorithms are not expected to benefit much when the sample size is sufficiently large. Similarly, due to the small sample sizes (100 and 200), to maintain a reasonable effective sample size, we limited the maximum number of categories to 12. In addition, we also analyzed a clinical trial document dataset that was collected by our group. We considered sample sizes of 100, 200, and 12 categories for this datasets to evaluate the performance of our method. 

\let\thefootnote\relax\footnotetext{Accessed: June 2023}

We used the BERT classification model and considered three augmentation methods to generate inputs, including 1) EDA: a combination of synonym replacement, random word insertions, swaps and deletions; 2) Cloud translation service (Google Translate): Translate to Chinese, Japanese, Korean and Hindi then back to English and; 3) Neural network based translation (OPUS-MT): Translate to French, Portuguese, Spanish and Italian then back to English. We also combined the three augmentation methods (denoted as ``Combiend''). Each original input text was augmented 4 times with each augmentation method, and this results in a total of 5 inputs including the original text. For comparison purposes, we also analyzed the data without augmentation (denoted as the baseline) and the traditional BERT model that treats each augmented text as if they are independent (denoted as the standard method). 

We chose 80\% of the data to train the model and used the remaining 20\% for evaluating the model performance. This process is repeated 25 times and the average accuracy of each method is reported. 

\subsection{Amazon Customer Reviews} 
\begin{figure}[h]
    \centering
    \includegraphics[scale=0.45]{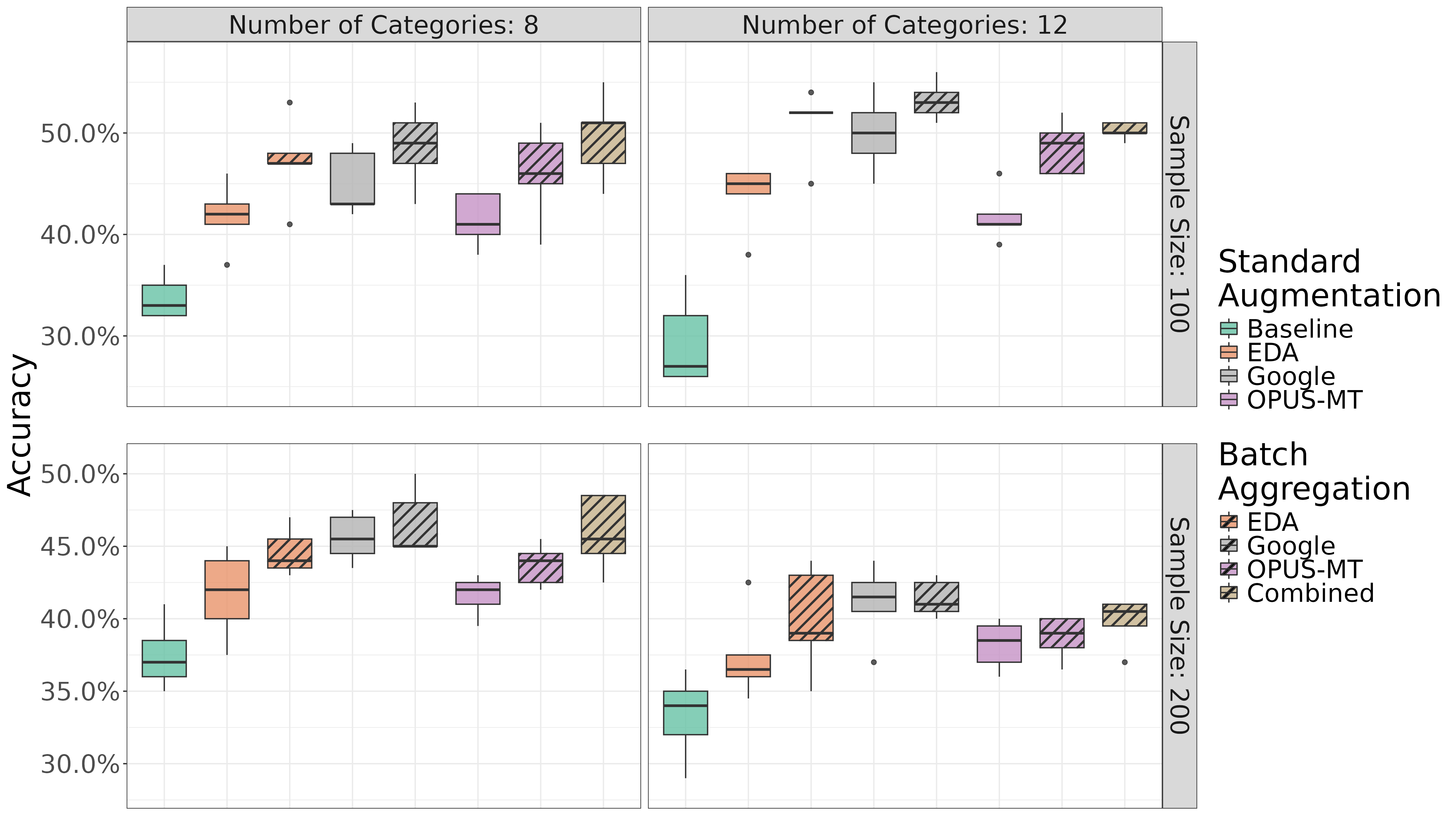}
    \caption{Average accuracies of Standard augmentation on the Amazon data set. ``Combined'' represents the combinations of EDA, Google and OPUS-MT.}
    \label{fig:amazon-result}
\end{figure}

This data set was released by Amazon for research purposes and contains over 130 million multi-language customer reviews and 46 classes. Its content includes information about the product which was purchased, review from the buyer, and their satisfaction score. Since the purpose of this study is to perform text classification, the customer review body (i.e., what the customers comments were about the purchase) was used as the input to classify the category of the corresponding product such as electronics, toy and apparel. After pre-processing the data and filtering out non English texts, we chose 12 classes which were the most populated. We also created an 8 class subset to allow for direct comparisons with benchmark datasets that do not have 12 classes and to evaluate model performance under different class configurations. This resulted in approximately 130,000 observations. The data set is summarized in \autoref{tab:amazon-dist}. Despite the original data has a large sample size, in our studies, we only consider a sample size of 100 and 200 for training as our as augmentation methods are mainly designed for small sample sizes. As mentioned in the previous section, for evaluation, we considered all combinations of sample size of 100 and 200, number of categories of 8 and 12 where possible. In terms of augmentation methods, we compared EDA, Google Translate and OPUS-MT. Results are summarized in \autoref{fig:amazon-result} and \autoref{tab:amazon-result}.  

As expected, the sample size increases and the number of categories decreases, the performance of all methods increase. Benchmark study on the open domain Amazon data set confirms that utilizing text augmentation does indeed improve accuracy, where the boost in performance is more obvious when the sample size is smaller. For the number of categories of 8 and 12, when the sample size is 100, the increase in accuracy between the baseline and the least performant standard augmentation method is 7.6\% and 12.4\% respectively. When the sample size is 200, the increase in accuracy for the two categories is 4.1\% and 3.9\% respectively. In addition, batch aggregation consistently results in higher accuracy when comparing with the standard method. This difference is most obvious when the sample size is 100 and the number of categories is 12, with improvements of 7.2\%, 3.2\%, and 6.8\% for EDA, Google, and OPUS-MT respectively. However, when sample size is 200, the increase in performance between the standard augmentation method and batch aggregation is becoming less significant. This is likely due augmentation being more efficient when $n$ is small. When considering a single augmentation method, Google back-translation consistently outperforms other methods, especially in smaller sample sizes. The highest accuracy when sample size is 100 is when all three augmentation methods are combined with batch aggregation, and when sample size is 200, the accuracy is somewhere between the three methods.

\begin{table}[htbp]
  \centering
  \begin{threeparttable}
  \renewcommand{\arraystretch}{1.2}
  \resizebox{1\textwidth}{!}{
  \begin{tabular}{>{\centering\arraybackslash}p{2.5cm}>{\centering\arraybackslash}p{3cm}c@{\hspace{1em}}ccc@{\hspace{1em}}cccc}
      \Thickrule
      \multirow{2}{*}{\parbox{2.5cm}{\centering Sample\\Size}} & 
      \multirow{2}{*}{\parbox{3cm}{\centering Number of\\Categories}} & 
      \multicolumn{1}{c}{Baseline} &
      \multicolumn{3}{c}{Standard} & 
      \multicolumn{4}{c}{Batch Aggregation (BAGG)} \\
      \cmidrule(lr){3-3} \cmidrule(lr){4-6} \cmidrule(lr){7-10}
      & & & EDA & Google & OPUS-MT & EDA & Google & OPUS-MT & Combined\tnote{a} \\
      \midrule
        100 & ~8 & 33.80\% & 41.80\% & \textbf{45.00\%} & 41.40\% & 47.20\% & 48.60\% & 46.00\% & \textit{49.60\%} \\
        200 & ~8 & 37.50\% & 41.70\% & \textbf{45.60\%} & 41.60\% & 44.60\% & \textit{46.60\%} & 43.70\% & 45.90\% \\
        100 & 12 & 29.40\% & 43.80\% & \textbf{50.00\%} & 41.80\% & 51.00\% & \textit{53.20\%} & 48.60\% & 50.20\% \\
        200 & 12 & 33.30\% & 37.40\% & \textbf{41.10\%} & 38.20\% & 39.90\% & \textit{41.40\%} & 38.70\% & 39.80\% \\
      \Thickrule
  \end{tabular}
  }
  \begin{tablenotes}
    \scriptsize
    \item[a] Combined represents the combinations of EDA, Google and OPUS-MT.
  \end{tablenotes}
  \end{threeparttable}
  \caption{Average accuracies of Standard augmentation and Batch Aggregation on the Amazon data set.  \textbf{Bold} and \textit{italic} values indicate the highest accuracy achieved for ``Standard augmentation'' and ``BAGG'', respectively.}
  \label{tab:amazon-result}
\end{table}

\subsection{The 20 Newsgroups Text Data} 
This data set is  well-known in the field of machine learning. It consists of 20 categories where each corresponds to a news topic such as Science: Electronics and Recreation: Baseball. The unique characteristics of this data set is that some topics may be very similar to each other, such as Computer: PC Hardware and Computer: Mac Hardware, and the number of observations in each category is also fairly consistent. This data set was chosen due to its wide span across different domains, from open-domain such as sport topics to very domain specific topics such as cryptography and computer hardware. In addition, the length of the input text is longer than the Amazon and LitCovid data sets. Details of the data set is summarized in \autoref{tab:newsgroup-dist}.

Similar to the previous results, BAGG outperforms the standard augmentation method and Google back-translation still yields the highest accuracy for most cases, except when sample size is 100 and number of categories is 8. Also, the increase in accuracy remains to be the most significant when sample size is 100 and number of categories is 12 at 67.4\% compared to the baseline at 45.6\% which is a 21.8\% increase. When sample size is larger, the improvement of batch aggregation becomes less significant, in fact the improvement of standard augmentation is also smaller compared to the baseline, this is expected since augmentation methods are designed for small data sets. Furthermore, when the number of categories increase, the overall improvement of augmentation decreases due to a smaller effective sample size. In terms of the combined method, it yielded the highest accuracy in 2 out of 8 cases, the rest are either between the three augmentation methods or slightly lower.

\begin{table}[htbp]
  \centering
  \begin{threeparttable}
  \renewcommand{\arraystretch}{1.2}
  \resizebox{1\textwidth}{!}{
  \begin{tabular}{>{\centering\arraybackslash}p{2.5cm}>{\centering\arraybackslash}p{3cm}c@{\hspace{1em}}ccc@{\hspace{1em}}cccc}
      \Thickrule
      \multirow{2}{*}{\parbox{2.5cm}{\centering Sample\\Size}} & 
      \multirow{2}{*}{\parbox{3cm}{\centering Number of\\Categories}} & 
      \multicolumn{1}{c}{Baseline} &
      \multicolumn{3}{c}{Standard} & 
      \multicolumn{4}{c}{Batch Aggregation} \\
      \cmidrule(lr){3-3} \cmidrule(lr){4-6} \cmidrule(lr){7-10}
      & & & EDA & Google & OPUS-MT & EDA & Google & OPUS-MT & Combined\tnote{a} \\
      \midrule
        100 & ~8 & 60.60\% & \textbf{72.20\%} & 69.60\% & 69.40\% & \textit{76.00\%} & 75.40\% & 74.40\% & 73.40\% \\
        200 & ~8 & 76.90\% & 78.20\% & \textbf{79.40\%} & 78.60\% & 80.50\% & \textit{81.70\%} & 80.90\% & 80.50\% \\
        100 & 12 & 45.60\% & 61.00\% & \textbf{63.40\%} & 63.00\% & 66.20\% & 67.40\% & 64.40\% & \textit{67.60\%} \\
        200 & 12 & 66.30\% & 71.50\% & \textbf{72.40\%} & 70.60\% & 72.10\% & 72.90\% & 72.50\% & \textit{73.50\%} \\
      \Thickrule
  \end{tabular}
  }
  \begin{tablenotes}
    \scriptsize
    \item[a] Combined represents the combinations of EDA, Google and OPUS-MT.
  \end{tablenotes}
  \end{threeparttable}
  \caption{Average accuracies of Standard augmentation on the Newsgroup data set.}
  \label{tab:newsgroup-results}
\end{table}

\begin{figure}[h]
	\centering
	\includegraphics[scale=0.45]{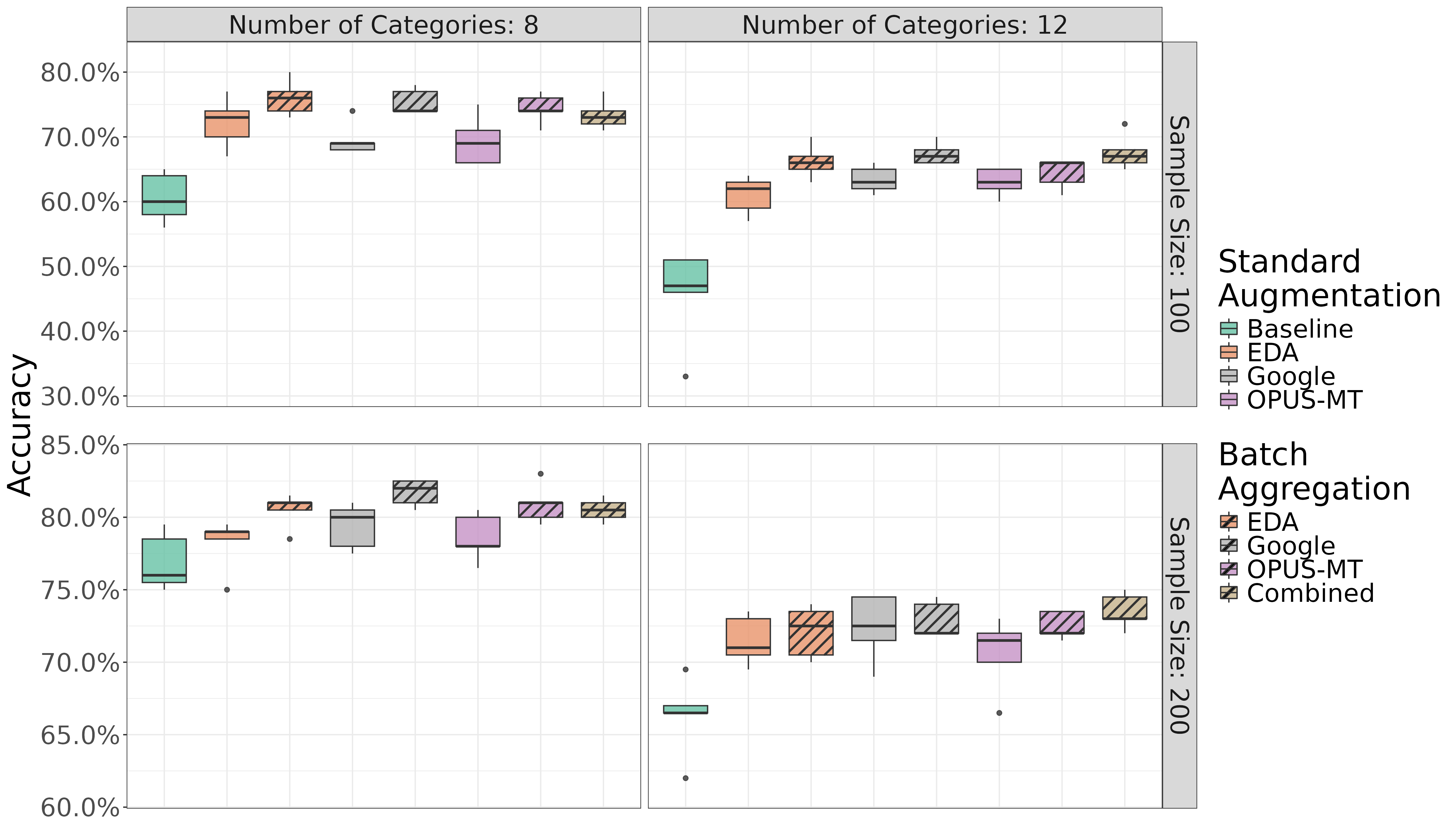}
	\caption{Average accuracies of Standard augmentation on the Newsgroup data set.}
	\label{fig:newsgroup-results}
\end{figure}

\subsection{LitCovid} 
LitCovid is a classification data set where one or more topics are assigned to a Covid-19 related PubMed article abstract and body. We chose this data set to examine the performance of batch aggregation with domain specific text. In total, there are approximately 22,500 observations and 7 topic labels: Treatment, Diagnosis, Prevention, Mechanism, Transmission, Epidemic Forecasting, and Case Report. Furthermore, the sample size of each category is highly imbalanced, there are 70 times more samples labelled as prevention in comparison to transmission (\autoref{tab:litcovid-dist}).

In the context of utilizing single augmentation methods, batch aggregation continues to outperform the standard augmentation method across all cases and batch aggregation remains to be the most effective when the sample size is 100. Here, the most performing method is OPUS-MT, where its accuracy is 17.8\% above the baseline. On average, batch aggregation is approximately 1.6\% more accurate than the standard method. Furthermore, in previous results, Google performed the best in all cases except with the Newsgroup dataset when the sample size was 100 and the number of categories was 8. However, in this data set, OPUS-MT consistently outperforms the other augmentation methods. The difference in result is likely due to the characteristics of this data set, domain specificity and imbalanced categories. When comparing with the 2 previous results, we observed that the Google back-translation performs better on general domain data sets. As for the combined method, LitCoivd is the only dataset where it consistently outperforms all single augmentation methods, this may be due to the fact that this is domain specific data set.

\begin{table}[htbp]
  \centering
  \begin{threeparttable}
  \renewcommand{\arraystretch}{1.2}
  \resizebox{1\textwidth}{!}{
  \begin{tabular}{>{\centering\arraybackslash}p{2.5cm}>{\centering\arraybackslash}p{3cm}c@{\hspace{1em}}ccc@{\hspace{1em}}cccc}
      \Thickrule
      \multirow{2}{*}{\parbox{2.5cm}{\centering Sample\\Size}} & 
      \multirow{2}{*}{\parbox{3cm}{\centering Number of\\Categories}} & 
      \multicolumn{1}{c}{Baseline} &
      \multicolumn{3}{c}{Standard} & 
      \multicolumn{4}{c}{Batch Aggregation} \\
      \cmidrule(lr){3-3} \cmidrule(lr){4-6} \cmidrule(lr){7-10}
      & & & EDA & Google & OPUS-MT & EDA & Google & OPUS-MT & Combined\tnote{a} \\
      \midrule
        100 & 7  & 69.20\% & \textbf{80.60\%} & 80.20\% & 79.60\% & 81.00\% & 80.40\% & 82.00\% & \textit{83.00\%} \\
        200 & 7  & 82.60\% & 82.70\% & 84.30\% & \textbf{84.50\%} & 86.30\% & 85.30\% & 86.50\% & \textit{87.00\%} \\
      \Thickrule
  \end{tabular}
  }
  \begin{tablenotes}
    \scriptsize
    \item[a] Combined represents the combinations of EDA, Google and OPUS-MT.
  \end{tablenotes}
  \end{threeparttable}
  \caption{Average accuracies of Standard augmentation on the LitCovid data set.}
  \label{tab:litcovid-result}
\end{table}

\begin{figure}[h]
	\centering
	\includegraphics[scale=0.45]{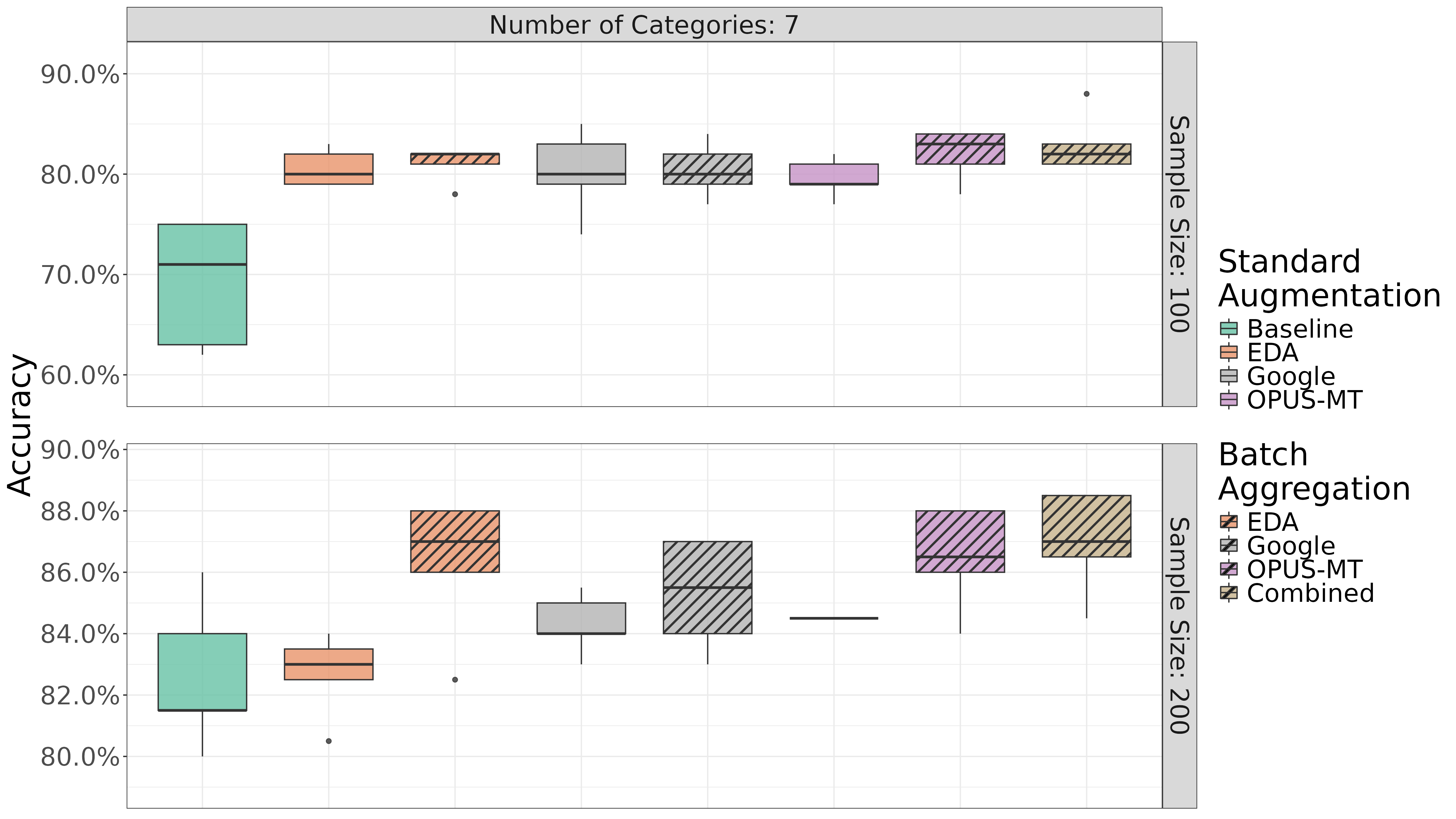}
	\caption{Average accuracies of Standard augmentation on the LitCovid data set.}
	\label{fig:result-litcovid}
\end{figure}

\subsection{Clinical Trial Documents}

Comprising 400 documents and designed for automating data analysis, this data intends to identify and recommend the most appropriate analytical methods for the clinical trials by extracting key information from each document, where each key information is constructed as a question-answer pair. Although this is not a classification data set, it is possible to utilize the question ``what is the statistical method used to analyze the primary endpoint?'' to predict the statistical method used for the study such as survival analysis. For this study, we excluded the documents that have multiple analytical methods used to analyze the primary endpoint. We also excluded those that employ a study-specific analysis, and thus only considered the following statistical models: ANCOVA, ANOVA, Chi-Squared Test, Cox Proportional-Hazards Model, Mixed-Effects Model, Cochran-Mantel-Haenszel Test, Fisher Exact Test, Log Rank Test, T-Test, Generalized Linear Model, Logistic Regression, and Generalized Estimating Equation. In total, there are 190 clinical trials documents with 12 categories (\autoref{tab:ct-dist}).

This benchmark study indicates some similar findings as the previous results, all augmentation methods shows an increase in accuracy and batch aggregation always outperforms the standard method. When considering only one augmentation method, the most significant increase in accuracy remains to be when the sample size is 100, where the best performing method is OPUS-MT at 72.8\% (29.2\% higher than the baseline). This improvement is significant and the largest we observed. In additional, like the LitCovid data set, clinical trials is also domain specific, our previous findings of Google being less performant on domain specific text still remains to be true. Also similar to the LitCovid data set, the combined method consistently outperform other single methods, except for when sample size is 100, at under 1\% lower than the least performing single model, Google back-translate. 

\begin{table}[htbp]
  \centering
  \begin{threeparttable}
  \renewcommand{\arraystretch}{1.2}
  \resizebox{1\textwidth}{!}{
  \begin{tabular}{>{\centering\arraybackslash}p{2.5cm}>{\centering\arraybackslash}p{3cm}c@{\hspace{1em}}ccc@{\hspace{1em}}cccc}
      \Thickrule
      \multirow{2}{*}{\parbox{2.5cm}{\centering Sample\\Size}} & 
      \multirow{2}{*}{\parbox{3cm}{\centering Number of\\Categories}} & 
      \multicolumn{1}{c}{Baseline} &
      \multicolumn{3}{c}{Standard} & 
      \multicolumn{4}{c}{Batch Aggregation} \\
      \cmidrule(lr){3-3} \cmidrule(lr){4-6} \cmidrule(lr){7-10}
      & & & EDA & Google & OPUS-MT & EDA & Google & OPUS-MT & Combined\tnote{a} \\
      \midrule
        100 & 12 & 43.60\% & 63.80\% & 64.80\% & \textbf{65.00\%} & 71.40\% & 70.20\% & \textit{72.80\%} & 69.40\% \\
        190 & 12 & 61.50\% & 72.50\% & 72.20\% & \textbf{73.70\%} & 79.60\% & 79.40\% & 79.60\% & \textit{81.20\%} \\
      \Thickrule
  \end{tabular}
  }
  \begin{tablenotes}
    \scriptsize
    \item[a] Combined represents the combinations of EDA, Google and OPUS-MT.
  \end{tablenotes}
  \end{threeparttable}
  \caption{Average accuracies of Standard augmentation on the Clinical Trials data set.}
  \label{tab:ct-result}
\end{table}

\begin{figure}[h]
    \centering
    \includegraphics[scale=0.45]{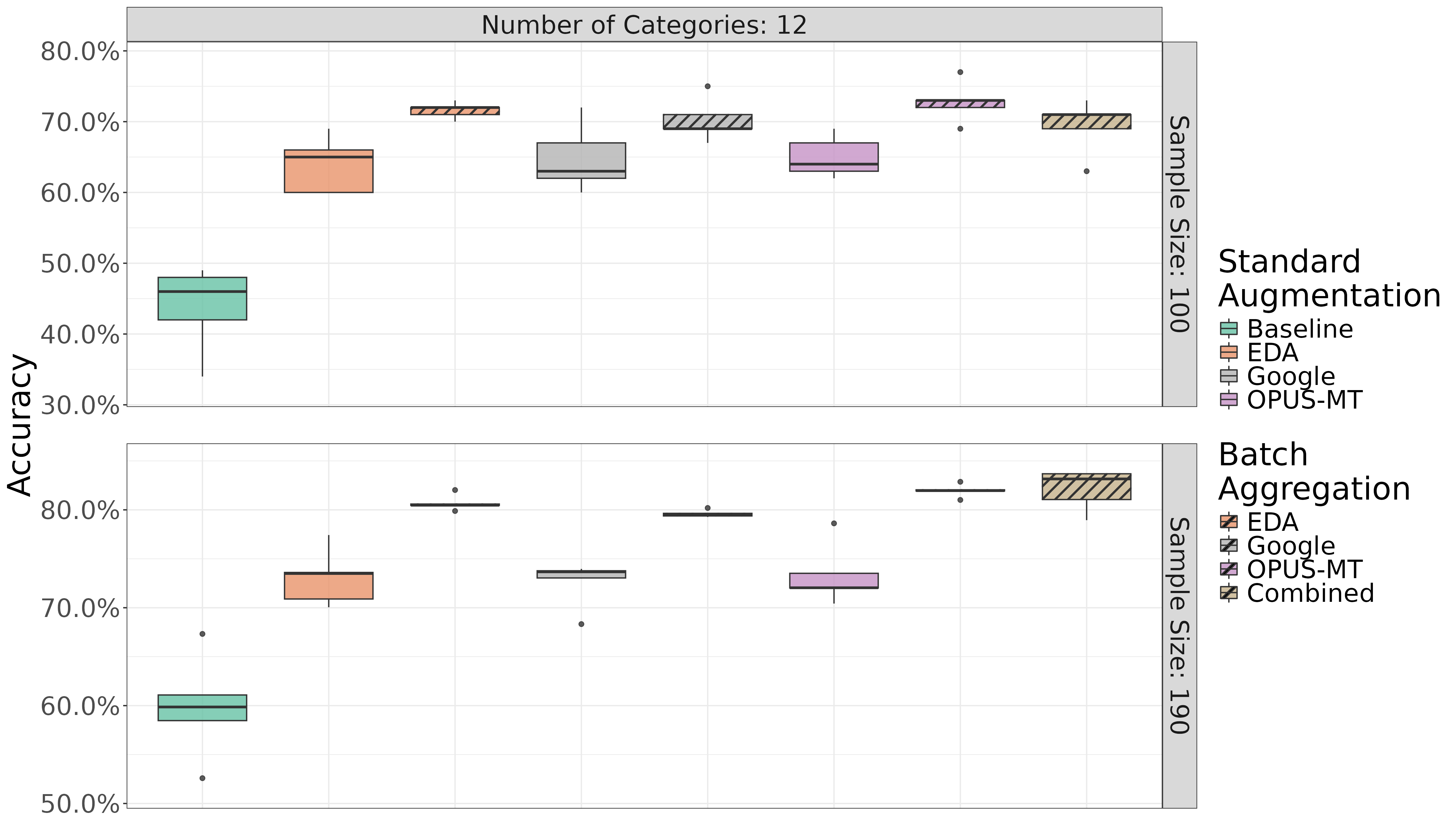}
    \caption{Average accuracies of Standard augmentation on the Clinical Trials data set.}
    \label{fig:ct-result}
\end{figure}
\section{Discussion} \label{sec:discussion}

We have found that regardless of augmentation methods used, augmentation benefits the model performance the most when the input sample size is small. It can result in an average of increase of 12.4\% when the input sample size is 100 in the benchmark datasets, whereas the improvement is only 9.1\% when the sample size in the training set is increased to 200 (Tables~\ref{tab:amazon-result}, \ref{tab:newsgroup-results}, \ref{tab:litcovid-result} and \ref{tab:ct-result}). In addition, there is an improvement of 13.7\% and 3.5\% when compared to the baseline method and those augmentation methods without explicitly considering correlations in the input text and this improvement in classification is observed consistently across augmentation methods and datasets regardless of whether it is open or close domain. In addition, rarely there are cases where BAGG performed worse than the standard augmentation.

Our study also suggests that no augmentation method is consistently superior. For example, the Google augmentation method works better for the Amazon and Newsgroup data set, especially when the sample size is smaller, whereas the OPUS-MT method is better for Clinical Trials data set. This is likely due to the difference in data characteristics and the domain specificity of each data set. Therefore, to increase the robustness of model performance, we also combined augmented texts from various methods and used the batch aggregation method to account for their relatedness among them. Although the combined batch aggregation method does not always provide the best performance, it seldom yeilds the worst performance. In fact, across all our findings, more than half of the cases showed that the combined method outperformed other approaches, showing its reliability and overall performance in various scenarios. In summary, we observed that batch aggregation outperforms all single augmentation methods when compared directly (i.e., EDA compared to EDA with batch aggregation). Google back-translate performs well on open domain tasks and OPUS-MT is more suitable for domain-specific tasks. Batch aggregation with combined augmentation methods tend to provide more robust results as compared to single augmentation method. 

Although our results suggests that BAGG is a promising method, providing increased predictive performance and reduced complexity when compared to standard methods, it also comes with its limitations. Our method relies heavily on the quality of the augmentation or the data itself. It is to be expected that batch aggregation will not be effective if the standard augmentation approach shows negative results since BAGG is simply based on the standard approach. Furthermore, despite our proposed batch aggregation method being flexible in modeling the contributions from each text input, we only considered an average layer. It would be worth investigating how to account for the quality of augmentation by incorporating more complex aggregation layers such as weighting or non-linear layers to the model. Additionally, our benchmark study considered a batch size of 5, as each input was augmented 4 times. Exploring the impact of larger batch sizes by increasing the number of augmented texts in the training set could be valuable. This can be achieved either through apply multiple augmentation methods or by using a single method multiple times. Moreover, a systematic approach to regularization and hyper-parameter tuning could be investigated to enhance model performance and generalization. 

\newpage
\addcontentsline{toc}{section}{References}
\newpage
\printbibliography[heading=bibintoc]

\newpage
\appendix
\section*{Appendix}

\setcounter{section}{0}
\setcounter{table}{0}
\renewcommand{\thetable}{A\arabic{table}}

\section{Benchmark Data Set Characteristics}

\subsection{Amazon}

\begin{table}[h]
\centering
\begin{tabular}{lccccccc}
\hline
\multirow{2}{*}{Label} & \multirow{2}{2cm}{\centering Sample\\Size} & \multicolumn{5}{c}{Token Length} & \\
\cline{3-7}
 &  & Mean & SD & Median & LQ & UQ & \\
\hline
appreal & 16,341 & 29.55 & 26.92 & 22.00 & 12.00 & 38.00 & \\
automotive & ~7,693 & 32.16 & 32.15 & 22.00 & 12.00 & 41.00 & \\
beauty & 12,392 & 33.35 & 32.33 & 24.00 & 12.00 & 43.00 & \\
drugstore & 12,019 & 32.93 & 32.09 & 23.00 & 12.00 & 43.00 & \\
electronics & ~6,339 & 37.97 & 40.45 & 26.00 & 13.00 & 48.00 & \\
home & 18,119 & 32.97 & 32.07 & 24.00 & 12.00 & 43.00 & \\
kitchen & 10,636 & 34.45 & 35.16 & 24.00 & 13.00 & 44.00 & \\
lawn and garden & ~7,500 & 35.07 & 36.92 & 24.00 & 13.00 & 44.00 & \\
pet products & ~7,250 & 37.43 & 36.30 & 26.00 & 14.00 & 49.00 & \\
sports & ~8,494 & 33.66 & 33.61 & 24.00 & 12.00 & 43.00 & \\
toy & ~8,986 & 31.23 & 29.49 & 23.00 & 12.00 & 41.00 & \\
wireless & 16,091 & 26.05 & 36.31 & 25.00 & 13.00 & 46.00 & \\
\hline
\end{tabular}
\caption{Amazon data set. LQ represents the Lower Quartile, and UQ represents the Upper Quartile.}
\label{tab:amazon-dist}
\end{table}     

\subsection{Newsgroup}

\begin{table}[h]
\centering
\begin{tabular}{lccccccc}
\hline
\multirow{2}{*}{Label} & \multirow{2}{2cm}{\centering Sample\\Size} & \multicolumn{5}{c}{Token Length} & \\
\cline{3-7}
 &  & Mean & SD & Median & LQ & UQ & \\
\hline
comp.graphics & 921 & 207.61 & 832.99 & ~67.00 & 40.00 & 129.00 & \\
comp.sys.ibm.pc.hardware & 936 & 131.42 & 247.27 & ~83.00 & 48.00 & 144.00 & \\
comp.sys.mac.hardware & 898 & 119.42 & 364.32 & ~75.00 & 45.00 & 125.00 & \\
comp.windows.x & 946 & 228.62 & 840.52 & ~83.50 & 45.00 & 157.75 & \\
rec.autos & 893 & 125.06 & 225.13 & ~80.00 & 44.00 & 143.00 & \\
rec.motorcycles & 912 & 108.80 & 207.69 & ~67.00 & 36.00 & 135.25 & \\
rec.sport.baseball & 913 & 135.72 & 176.77 & ~79.00 & 40.00 & 158.00 & \\
rec.sport.hockey & 937 & 196.48 & 508.47 & ~93.00 & 48.00 & 183.00 & \\
sci.crypt & 928 & 240.58 & 598.52 & 115.00 & 56.75 & 218.00 & \\
sci.electronics & 930 & 126.61 & 398.92 & ~84.00 & 47.00 & 140.00 & \\
sci.med & 943 & 209.95 & 477.04 & ~97.00 & 51.00 & 189.00 & \\
sci.space & 932 & 200.89 & 508.25 & ~92.00 & 48.00 & 176.25 & \\
\hline
\end{tabular}
\caption{Newsgroups data set. See https://huggingface.co/datasets/rungalileo/20\_Newsgroups\_Fixed/ for details.}
\label{tab:newsgroup-dist}
\end{table}

\newpage

\subsection{LitCovid}

\begin{table}[h]
\centering
\begin{tabular}{lccccccc}
\hline
\multirow{2}{*}{Label} & \multirow{2}{2cm}{\centering Sample\\Size} & \multicolumn{5}{c}{Token Length} & \\
\cline{3-7}
 &  & Mean & SD & Median & LQ & UQ & \\
\hline
case report & 2,742 & 141.58 & ~70.22 & 131.00 & 88.00 & 188.00 & \\
diagnosis & 3,243 & 216.24 & ~85.89 & 221.00 & 162.00 & 259.00 & \\
epidemic forecasting & ~~~265 & 176.06 & ~78.55 & 171.00 & 118.00 & 224.00 & \\
mechanism & ~~~808 & 180.56 & ~79.75 & 179.00 & 126.75 & 230.00 & \\
prevention & 12,049 & 198.06 & ~87.42 & 198.00 & 137.00 & 251.00 & \\
transmission & ~~~172 & 170.62 & ~85.97 & 178.00 & 97.00 & 220.00 & \\
treatment & ~3,268 & 210.60 & 120.47 & 198.00 & 140.00 & 251.00 & \\
\hline
\end{tabular}
\caption{LitCovid data set}
\label{tab:litcovid-dist}
\end{table}

\subsection{Clinical Trials}

\begin{table}[h]
\centering
\begin{tabular}{lccccccc}
\hline
\multirow{2}{*}{Label} & \multirow{2}{2cm}{\centering Sample\\Size} & \multicolumn{5}{c}{Token Length} & \\
\cline{3-7}
 &  & Mean & SD & Median & LQ & UQ & \\
\hline
ANCOVA & 41 & ~80.07 & 50.62 & ~76.00 & 51.00 & 108.00 & \\
ANOVA & 23 & ~76.22 & 39.42 & ~64.00 & 47.50 & ~97.50 & \\
Chi-squared test & ~7 & 101.14 & 48.28 & ~82.00 & 64.50 & 138.50 & \\
Cochran-Mantel-Haenszel test & 10 & ~95.30 & 57.54 & ~68.50 & 56.25 & 137.00 & \\
Cox proportional-hazards model & 28 & ~88.14 & 46.40 & ~82.50 & 46.00 & 115.75 & \\
Fisher exact test & 10 & ~63.70 & 30.82 & ~67.00 & 39.75 & ~88.25 & \\
Generalized estimating equation & ~4 & 133.50 & 76.29 & 115.50 & 95.75 & 153.25 & \\
Generalized linear model & ~7 & 103.71 & 75.01 & ~70.00 & 48.00 & 143.50 & \\
Log rank test & 29 & ~59.10 & 29.34 & ~59.00 & 29.00 & ~79.00 & \\
Logistic regression & ~7 & ~72.43 & 43.95 & ~69.00 & 33.00 & 105.00 & \\
Mixed-effects model & 15 & 144.00 & 60.55 & ~85.00 & 78.50 & 158.00 & \\
t-test & ~9 & ~79.78 & 35.83 & ~92.00 & 52.00 & ~97.00 & \\
\hline
\end{tabular}
\caption{Clinical Trials data set}
\label{tab:ct-dist}
\end{table}

\newpage

\setcounter{section}{18} % S is the 19th letter
\setcounter{figure}{0} % reset figure counter
\renewcommand{\thefigure}{S\arabic{figure}}
\section{Supplementary Material}
\begin{figure}[h]
	\centering
	\includegraphics[scale=0.35]{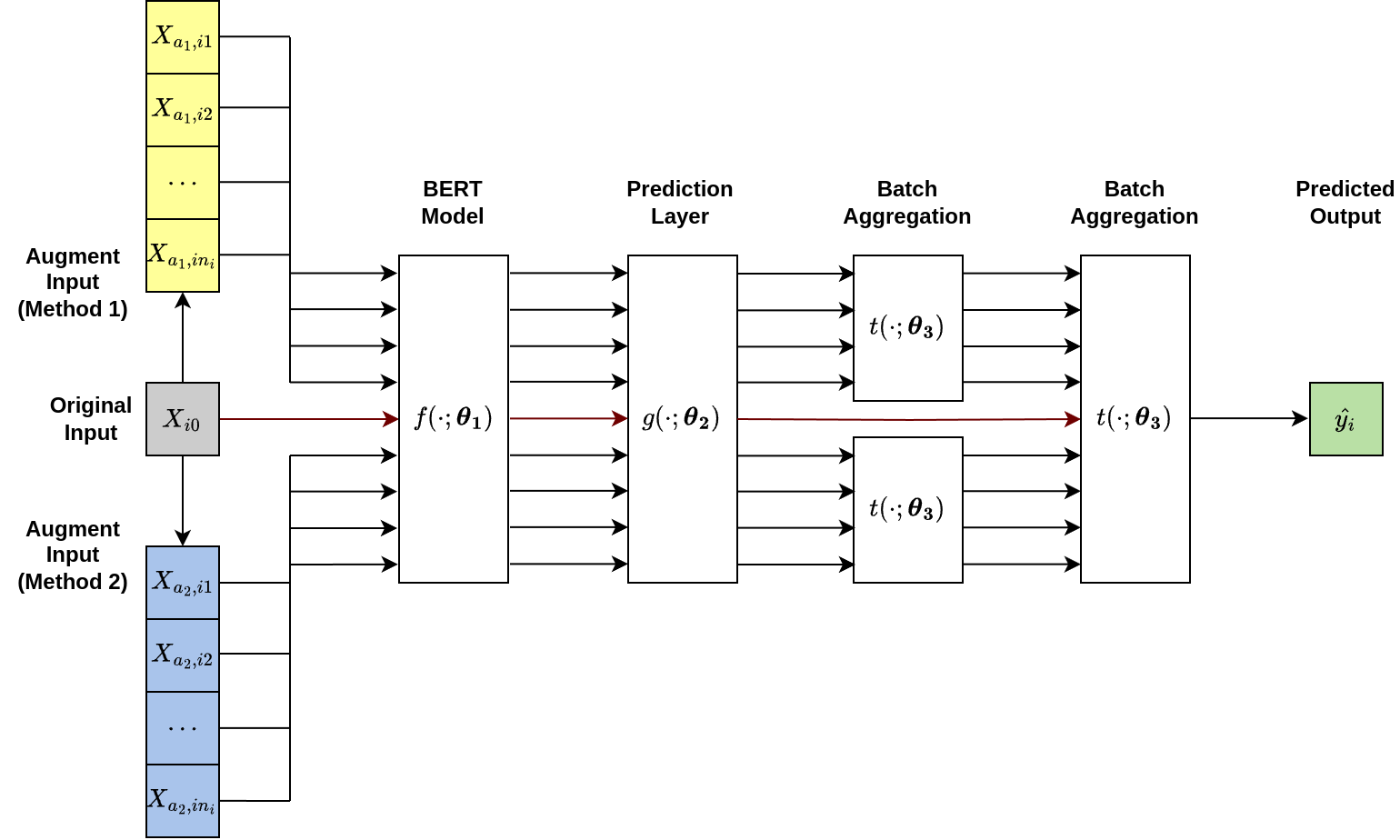}
	\caption{Batch Aggregation with 2 pooling layers}
	\label{fig:batch-aggregation-2pool}
\end{figure}

\end{document}